\documentclass[runningheads,orivec]{llncs}
\usepackage[T1]{fontenc}
\usepackage{graphicx, subcaption, tabularx}
\usepackage{caption, enumerate, url, float, url, hyperref}
\usepackage{amsfonts, amsmath, amssymb}
\usepackage{algorithm}
\usepackage{algorithmic}
\usepackage[table]{xcolor}
\begin{document}
\title{Sparse Orthogonal Regression Technique: A Spectral Framework for Equation Discovery, Approximation, and Integration}
\titlerunning{Sparse Orthogonal Regression Technique}
%
\author{Sabin Roman\inst{1} \and
Ljup{\v{c}}o Todorovski\inst{2, 1}\and
Sa{\v{s}}o D{\v{z}}eroski \inst{1}}
%
%
\institute{Department of Knowledge Technologies, Jo\v{z}ef Stefan Institute\\
\email{sabin.roman@ijs.si}\\
\and
Faculty of Mathematics and Physics, University of Ljubljana}
\maketitle              
\begin{abstract}
We develop the Sparse Orthogonal Regression Technique (SORT), a sparse spectral framework for learning orthonormal-basis expansions from noisy and irregularly sampled data. SORT estimates expansion coefficients directly from observations using \(L^1\)-regularized regression, avoiding explicit quadrature or analytic inner-product evaluation. The central application is data-driven discovery of ordinary differential equations: vector fields are represented in chosen orthogonal bases and learned as sparse coefficient expansions. This provides a complementary route to symbolic regression, grammar-based discovery, and SINDy-style sparse identification by first recovering a compact spectral representation, which can later guide searches for simpler analytic forms. Across the dynamical-system experiments, SORT matches or improves upon library-based sparse-regression baselines when the basis is well adapted to the problem, and shows more stable degradation under sparse sampling, noisy derivative estimates, and representation mismatch. Specific examples illustrate why this representation is useful: if a finite library misses the problem-specific nonlinearity, the resulting model can fail. SORT is not immune to mismatch, but it shifts the problem away from brittle selection among generic terms to basis design adapted to the problem domain. The experiments also show that dominant low-order coefficients persist as model order increases, supporting order-consistent model growth. Beyond equation discovery, the same learned expansion supports nonlinear approximation and estimation of complex, high-dimensional integrals by coefficient readout. Overall, SORT provides a reusable intermediate representation for system identification, approximation, and integration, while making basis design an explicit part of the scientific modeling problem.
\end{abstract}
\section{Introduction}

Many problems in scientific computing and machine learning require learning an unknown function from data and reusing that representation for downstream tasks. Examples include surrogate modelling for expensive simulations \cite{steinmann2025scenario,blatman2011adaptive,doostan2011nonadapted}, numerical integration from sampled observations \cite{boyd2001chebyshev,tran2022analysis}, and data-driven modelling or discovery of governing equations from time-series data \cite{brunton2016discovering,rudy2017data,roman2025maximum,roman2025toward}. Although these tasks are often treated separately, they share a common requirement: reconstructing a functional relationship from finite, noisy, and often irregularly sampled data in a stable and reusable form.

This paper develops the \emph{Sparse Orthogonal Regression Technique} (SORT) as a practical framework for estimating sparse expansions in prescribed orthonormal bases from sampled observations. SORT combines two standard ingredients: orthonormal basis representations and sparsity-promoting regression. We therefore do not claim algorithmic novelty in sparse regression on orthogonal features. The contribution is application-level and representational: SORT treats the learned sparse orthonormal expansion as a reusable object for data-driven dynamical-system reconstruction, numerical integration by coefficient readout, and nonlinear function approximation with order-consistent model growth.

The central application is dynamical-system reconstruction. Sparse identification methods such as SINDy \cite{brunton2016discovering,rudy2017data} learn governing equations by regressing derivatives onto finite libraries of candidate nonlinearities. This is effective when the correct terms are present in the library, but recovery can become ill-conditioned or brittle when the true dynamics are not sparse in the chosen dictionary \cite{feng2026illconditioning}. SORT provides a complementary route: each component of the vector field is represented in a fixed orthonormal basis and estimated through sparse regression. The output is not necessarily a compact symbolic expression, but a sparse spectral representation of the vector field. This use of orthonormal expansions for equation discovery remains comparatively underexplored relative to library-based sparse identification and symbolic regression \cite{kulkarni2020sparse}.

A second application is numerical integration from data. In an orthonormal expansion, coefficients are inner products between the function and basis elements. If a basis direction is chosen to represent an integration functional, then the corresponding coefficient gives the integral, up to a known normalization factor. Classical spectral methods usually obtain these coefficients by quadrature, exact projection, or structured transforms. SORT instead estimates them from scattered or noisy point samples by sparse regression and then evaluates the desired integral by coefficient readout. This gives a direct route from sampled data to integral estimates in settings where explicit quadrature is difficult, high-dimensional, or unavailable.

A third application is order-consistent model growth. In an orthonormal expansion, increasing the truncation order does not change the meaning of lower-order basis functions. Low-order coefficients describe coarse structure, while higher-order coefficients refine the approximation. In finite, noisy, and irregularly sampled data this hierarchy is only approximate, because empirical orthogonality and coefficient estimates depend on sampling and regularization. Nevertheless, coefficient persistence across increasing truncation orders provides a practical diagnostic of stable learning. This hierarchy is standard in numerical analysis but remains underused in machine-learning workflows, where increasing model complexity often changes the full parameterization rather than refining an existing coefficient structure. In this sense, SORT also provides a natural way to compare learned models ordered by truncation level, sparsity pattern, coefficient persistence, and predictive accuracy within a common representation \cite{roman2026approximating}.

We evaluate SORT across dynamical-system identification, numerical integration, and nonlinear approximation tasks. The experiments compare against dense least-squares, kernel-based, and library-based sparse-regression baselines under noise and subsampling. Across these settings, SORT is used not as a new sparse-regression algorithm, but as a sparse spectral workflow: it estimates orthonormal coefficients from data and reuses them for prediction, integration, vector-field reconstruction, and model refinement.

\section{Background and related work}
\label{sec:related}

SORT builds on several established areas: spectral approximation, sparse recovery in orthogonal systems, polynomial chaos expansions, sparse regression, and data-driven equation discovery. These areas share many mathematical ingredients, but they differ in their home fields, target applications, and intended outputs.

Classical spectral approximation represents functions in global orthogonal bases such as Fourier, Chebyshev, or Legendre systems \cite{boyd2001chebyshev}. Its home field is numerical analysis, where the main goals are accurate approximation of smooth functions and efficient evaluation of derivatives, integrals, or differential operators. Coefficients are typically computed by projection, quadrature, interpolation, or structured transforms. SORT keeps the spectral representation, but estimates coefficients from finite, noisy, and irregularly sampled data using sparse regression.

Sparse recovery in orthogonal polynomial bases has been studied extensively in approximation theory, compressed sensing, and uncertainty quantification. Early work showed that sparse Legendre expansions can be recovered from random samples via $\ell_1$ minimization \cite{rauhut2012legendre}, with later extensions to Chebyshev, Hermite, and other orthogonal systems, including analyses of sampling, conditioning, and high-dimensional stability \cite{hampton2015compressive,tran2022analysis,gilbert2020sparse}. This literature provides the mathematical basis for recovering compressible orthogonal expansions from limited samples. SORT uses this recovery principle as a modelling component, but does not propose new sparse-recovery theory.

Sparse Polynomial Chaos Expansion (PCE) methods are particularly close in methodology \cite{luthen2021sparse}. PCE represents the output of a model with uncertain inputs as an expansion in orthogonal polynomials matched to the input probability distribution. Sparse PCE combines this representation with sparse regression, adaptive truncation, or weighted penalties to control the growth of high-dimensional polynomial bases \cite{blatman2011adaptive,doostan2011nonadapted,peng2014weighted,jakeman2015enhancing}. Its main use is surrogate modelling and uncertainty quantification: once the expansion is learned, coefficients are used to compute statistical quantities such as means, variances, sensitivity indices, and other moments under a prescribed input distribution. SORT overlaps with sparse PCE at the level of orthogonal bases and sparse coefficient estimation, but its goal is different. It is geared toward dynamical-system reconstruction, direct integral estimation from sampled functions, and order-consistent approximation, rather than primarily toward uncertainty propagation or statistical post-processing.

At the algorithmic level, SORT relies on classical sparsity-promoting regression. Convex $\ell_1$-regularized regression \cite{tibshirani1996lasso}, path-following algorithms \cite{efron2004lars}, and compressed sensing theory \cite{candes2006robust} provide the tools and recovery intuition. Related orthogonal forward regression methods also use orthogonalization to reduce collinearity and improve sparse model selection \cite{chen2004sparse,billings2007model,tang2018stability,guo2016ultra,hong2015elastic}. These methods typically begin with a finite candidate library and use orthogonalization or regularization to select a stable subset. SORT instead begins with a prescribed orthonormal coordinate system, so increasing the truncation order refines the representation while preserving the meaning of lower-order coefficients.

Data-driven equation discovery provides the closest application-level comparison. SINDy and related methods identify governing equations by sparse regression over nonlinear libraries constructed from the observed variables \cite{brunton2016discovering,rudy2017data}. Symbolic regression searches more broadly over expression trees, grammars, or probabilistic symbolic structures \cite{guimera2020bayesian,brence2021probabilistic}. These approaches can recover compact, human-readable equations when the library or grammar matches the true system. SORT targets the same broad scientific-discovery setting but returns a sparse spectral vector-field model rather than a symbolic formula. This can be advantageous when the correct analytic building blocks are unknown, poorly matched by standard libraries, or too expensive to search over directly.

Overall, SORT is best viewed as a task-oriented recombination of established tools rather than as a new optimization method. This is analogous to the way SINDy made sparse regression useful for equation discovery by pairing it with nonlinear libraries. SORT instead places sparse orthonormal coefficients at the centre of the workflow: the same representation can define a spectral vector field, evaluate integrals through linear functionals, approximate nonlinear functions, and compare models across truncation orders.

\subsection{Mathematical notation}
\label{sec:sort-formulation}

Let \((\Omega,\mu)\) be the domain and measure with respect to which the basis is orthonormal, with inner product \(\langle f,g\rangle=\int_{\Omega}f(x)g(x)\,d\mu(x)\). Let \(\{\phi_k\}_{k\ge 0}\) be an orthonormal basis, so that \(\langle \phi_j,\phi_k\rangle=\delta_{jk}\). The spectral coefficients of \(f\) are \(c_k=\langle f,\phi_k\rangle\), and a finite truncation over an index set \(\Lambda_K\) is
\[
    f_K(x)=\sum_{k\in\Lambda_K} c_k\phi_k(x).
\]
This representation is useful when the coefficient sequence decays rapidly. For example, if \(f\) is \(2\pi\)-periodic and of bounded variation \(V\), then \(|\hat f_k|\le V/(2\pi |k|)\) for \(k\neq 0\), and the \(K\)-mode Fourier approximation satisfies \(\|f-S_Kf\|_{L^2}\le V/\sqrt{\pi K}\) \cite{stein2011fourier}. Thus, many smooth or compressible functions can be represented accurately by relatively few coefficients.

In data-driven settings, the inner products \(c_k=\langle f,\phi_k\rangle\) are not available. Instead, we observe samples \(y_i=f(x_i)+\varepsilon_i\), \(i=1,\dots,N\). SORT forms the design matrix \(\Phi_{ik}=\phi_k(x_i)\) and estimates the coefficient vector by
\[
    \hat c \in \arg\min_{u\in\mathbb{R}^{K}}
    \frac{1}{2N}\|y-\Phi u\|_2^2+\lambda\|u\|_1 .
\]
The learned approximation is \(\hat f(x)=\sum_{k\in\Lambda_K}\hat c_k\phi_k(x)\). The regularization parameter \(\lambda\) suppresses weak or noise-supported modes and is selected empirically, for example by validation error or cross-validation, with preference for the sparsest stable model.

In the ideal empirically orthonormal case, if \((1/N)\Phi^\top\Phi=I_K\) and \(\|(1/N)\Phi^\top\varepsilon\|_\infty\le\lambda\), standard LASSO arguments imply
\[
    \|\hat c-c\|_\infty\le 2\lambda
\]
\cite{tibshirani1996lasso,hastie2015statistical}. This bound is used only as background intuition: with irregular finite samples, empirical orthogonality is approximate, so coefficient stability must be assessed numerically.

For numerical integration, the relevant fact is that integrals are linear functionals. If \(\mathcal{I}(f)=\int_{\Omega}f(x)\,d\nu(x)\) has Riesz representer \(r_{\mathcal{I}}\in L^2(\Omega,\mu)\), then \(\mathcal{I}(f)=\langle f,r_{\mathcal{I}}\rangle\). If the normalized direction \(\phi_{\mathcal{I}}=r_{\mathcal{I}}/\|r_{\mathcal{I}}\|\) is included as a basis element, its coefficient satisfies
\[
    c_{\mathcal{I}}=\langle f,\phi_{\mathcal{I}}\rangle,
    \qquad
    \mathcal{I}(f)=\|r_{\mathcal{I}}\|\,c_{\mathcal{I}} .
\]
SORT estimates \(c_{\mathcal{I}}\) from point samples by sparse regression, after which the integral is obtained by coefficient readout. In the simplest case \(d\nu=d\mu\) on a finite-measure domain, \(r_{\mathcal{I}}(x)=1\), so the integral is proportional to the coefficient of the normalized constant basis function.

For equation discovery, consider an autonomous system \(\dot x(t)=f(x(t))\). SORT represents each vector-field component as
\[
    f_j(x)=\sum_{k\in\Lambda_K} c_{jk}\phi_k(x),
    \qquad j=1,\dots,d,
\]
and estimates the coefficients from samples of \((x(t),\dot x(t))\). The result is a sparse spectral vector-field model: explicit and reusable, but not necessarily a compact symbolic expression.

Basis choice should reflect the domain and regularity of the target function: Fourier bases for periodic or oscillatory structure, Legendre or Chebyshev bases on bounded intervals, Hermite or Laguerre families on unbounded or semi-infinite domains, and transformed rational bases when mapping physical domains to canonical intervals. In low dimensions, total-degree truncation is often sufficient; in higher dimensions, hyperbolic-cross or anisotropic truncation helps control basis growth. For \(N\) samples and \(K\) basis functions, constructing the design matrix costs \(O(NK)\), and coordinate-descent LASSO solvers scale with repeated passes over this matrix. The main bottleneck is the growth of \(K\) with dimension and truncation order. SORT does not remove the curse of dimensionality, and severe basis mismatch can degrade performance. When compact human-readable formulas are required, symbolic regression or library-based sparse identification may remain preferable.

\begin{figure}[!t]
  \centering
  \subfloat[Dense-trajectory RMSE vs sampling interval.]{
    \includegraphics[width=0.45\linewidth, height=0.2\textheight]{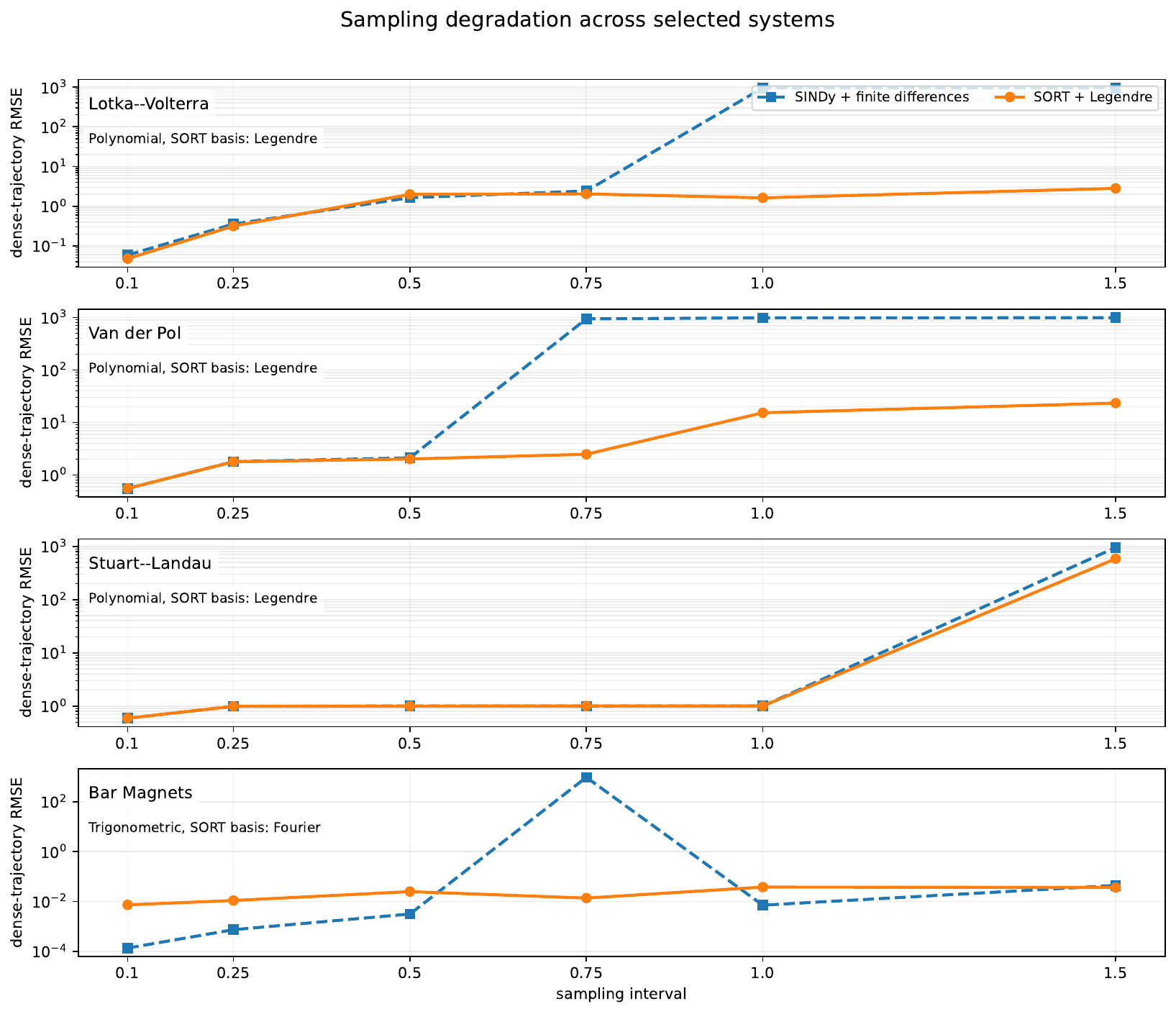}
    \label{fig:lv_sampling_rmse}
  }\hfill
  \subfloat[Representative rollouts at \(\Delta t_{\mathrm{sample}}=0.75\).]{
    \includegraphics[width=0.45\linewidth, height=0.2\textheight]{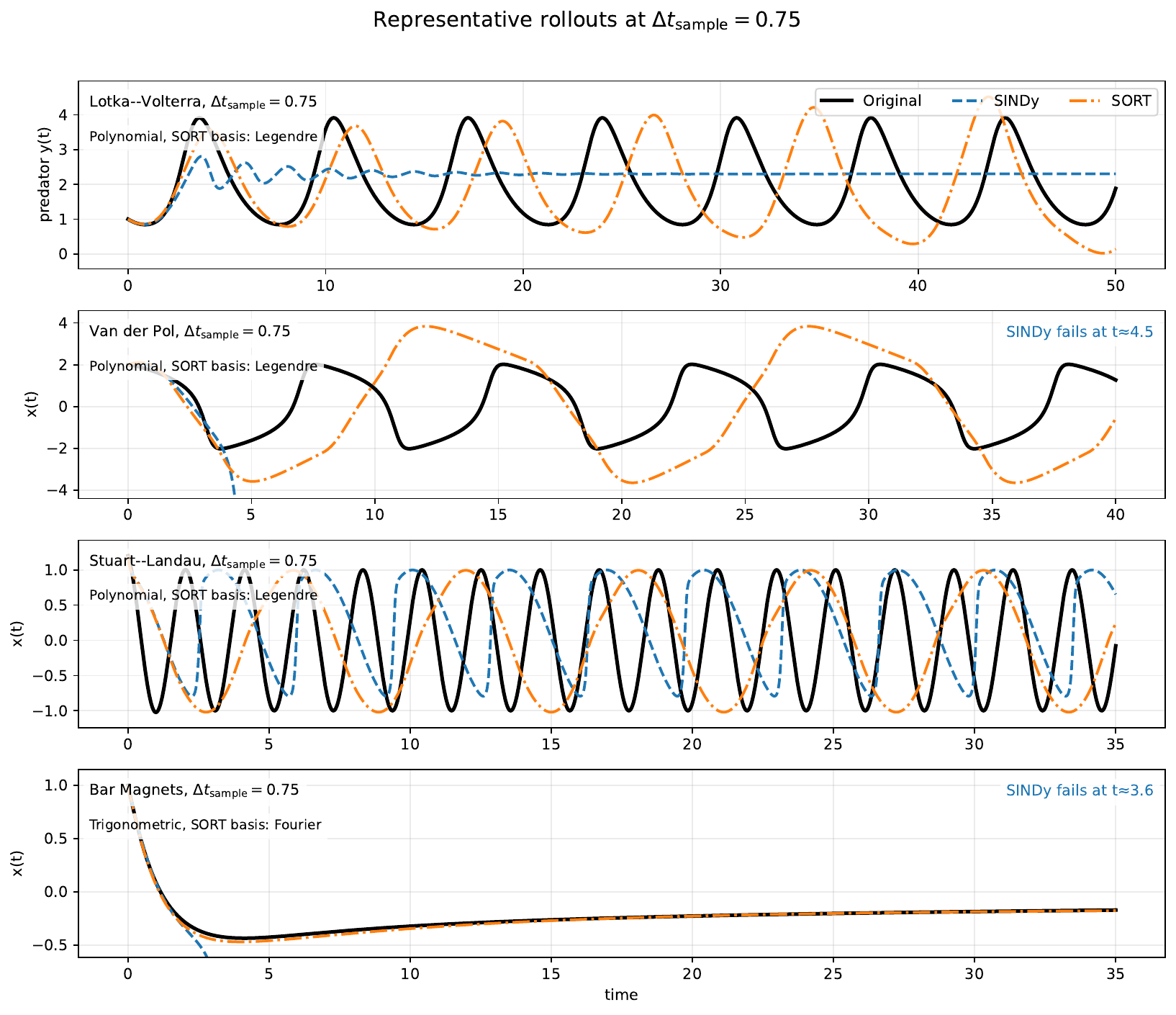}
    \label{fig:lv_sampling_predator}
  }
  \caption{\textbf{Sampling degradation under finite-difference derivative estimates.}
  Panel (a) reports dense-rollout RMSE as the sampling interval increases; panel (b) shows representative rollouts at \(\Delta t_{\mathrm{sample}}=0.75\).
  Both methods use the same sampled states and derivative estimates.
  Polynomial systems use degree-three SINDy and degree-three Legendre SORT; Bar Magnets uses Fourier features.
  The main effect is not better fine-sampling accuracy, but slower and less catastrophic degradation under coarse sampling.}
  \label{fig:lv_sampling_degradation}
\end{figure}

\section{Applications}
\label{sec:applications}

We evaluate SORT on data-driven system reconstruction, numerical integration by coefficient readout, and nonlinear approximation, focusing on data-limited regimes where representation affects stability.

\subsection{Equation discovery from time-series data}
\label{sec:equation-discovery}

We first study autonomous vector-field recovery from sampled trajectories. For each sampling interval, SINDy and SORT are trained on the same states, with derivatives estimated by finite differences and no smoothing. The comparison therefore isolates the learned representation, not the derivative estimator.

The examples are selected from Dynobench~\cite{omejc2024probabilistic}, restricted here to polynomial and smooth trigonometric systems. Polynomial cases use a degree-three SINDy library and a matching degree-three Legendre basis; the trigonometric case uses Fourier features. We exclude the coupled phase oscillator because it is non-autonomous and would require time or forcing terms. Bacterial respiration, predator--prey, glider, and shear flow contain rational, singular, or ratio-trigonometric terms such as \(1/x\) or \(\cot(y)\), which would require task-specific rational, transformed, or mixed bases. Lorenz is excluded for a different reason: its standard polynomial form gives a polynomial SINDy library a representational advantage reflecting our usual notation for the equations, not any principle that natural dynamics are polynomial. Thus, basis design is part of the scientific modeling problem, not an implementation detail, as in the use of Bessel functions for problems with radial or cylindrical symmetry. Different bases may be universal in principle, but under finite data, noise, and degraded derivative estimates, the coordinate system matters: some representations make the dynamics sparse and learnable, while others make the same system brittle or unnecessarily high-dimensional.

Figure~\ref{fig:lv_sampling_degradation} shows that both methods behave similarly when sampling is fine and derivatives are reliable. As sampling becomes coarse, Lotka--Volterra and Van der Pol show abrupt SINDy rollout failure, while the orthogonal sparse model remains bounded and degrades more gradually. Stuart--Landau is a neutral case, and Bar Magnets tests the Fourier setting. Therefore, with a suitable basis, sparse orthogonal regression is competitive with a library-based sparse model, while showing better behavior under degraded derivative information. Chaotic systems also require attractor-level evaluation rather than long-horizon pointwise trajectory error; we illustrate this issue separately below.

We next evaluate robustness to noise and representation mismatch using two three-dimensional cyclic systems. In these experiments, the trajectories are corrupted with additive Gaussian noise and then subsampled. Because the data are noisy, derivatives are estimated by applying a Savitzky--Golay filter to each observed coordinate and differentiating the smoothed signal. The same smoothing, derivative estimates, noise model, subsampling strategy, rollout horizon, and evaluation metric are used for SORT and the SINDy-style baseline. This second experiment therefore tests a different failure mode from the Lotka--Volterra sampling study: not coarse finite-difference degradation alone, but the interaction of noise, smoothing, and basis mismatch.

\begin{figure}[!t]
  \centering
  \subfloat[Thomas: RMSE vs train fraction.]{
    \includegraphics[width=0.45\linewidth]{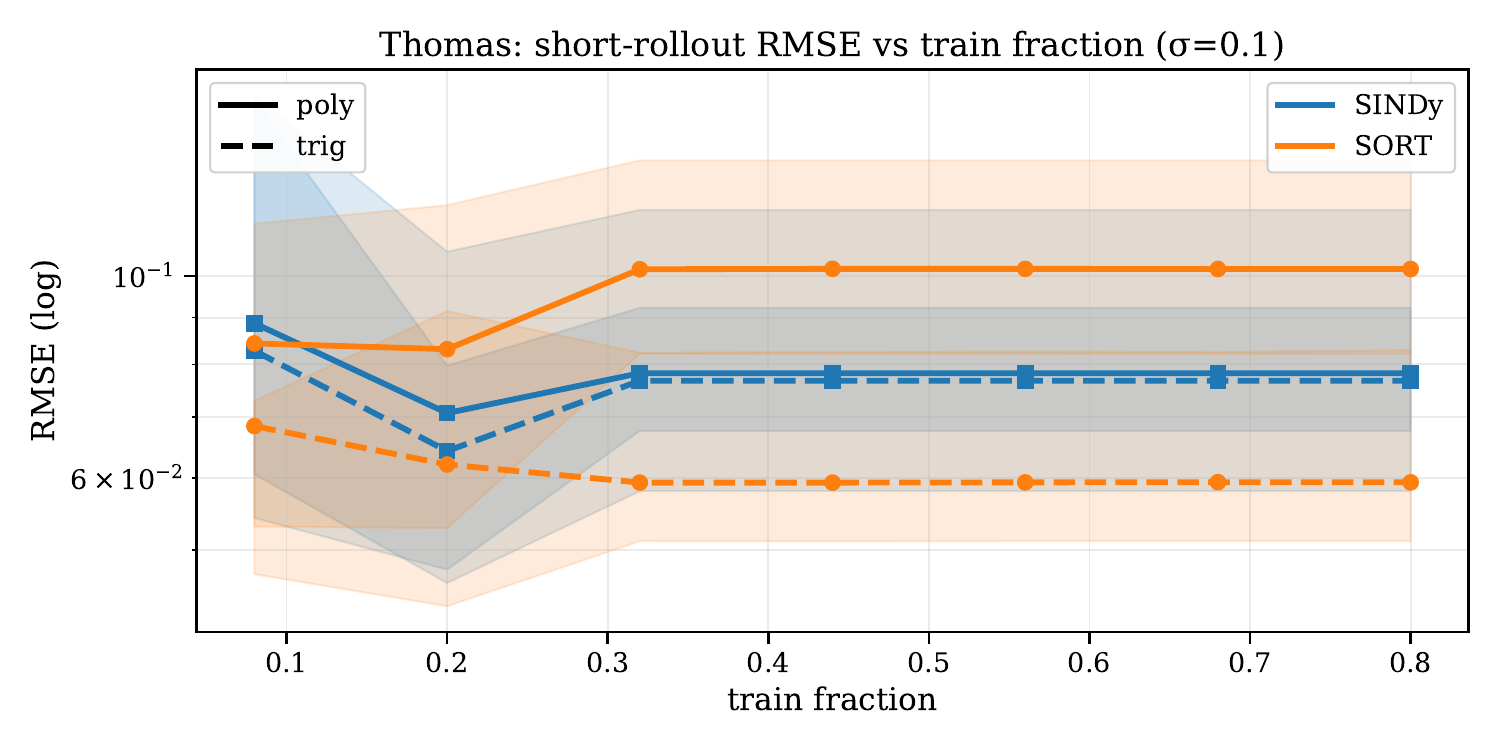}
    \label{fig:thomas_frac}
  }\hfill
  \subfloat[Thomas: RMSE vs noise.]{
    \includegraphics[width=0.45\linewidth]{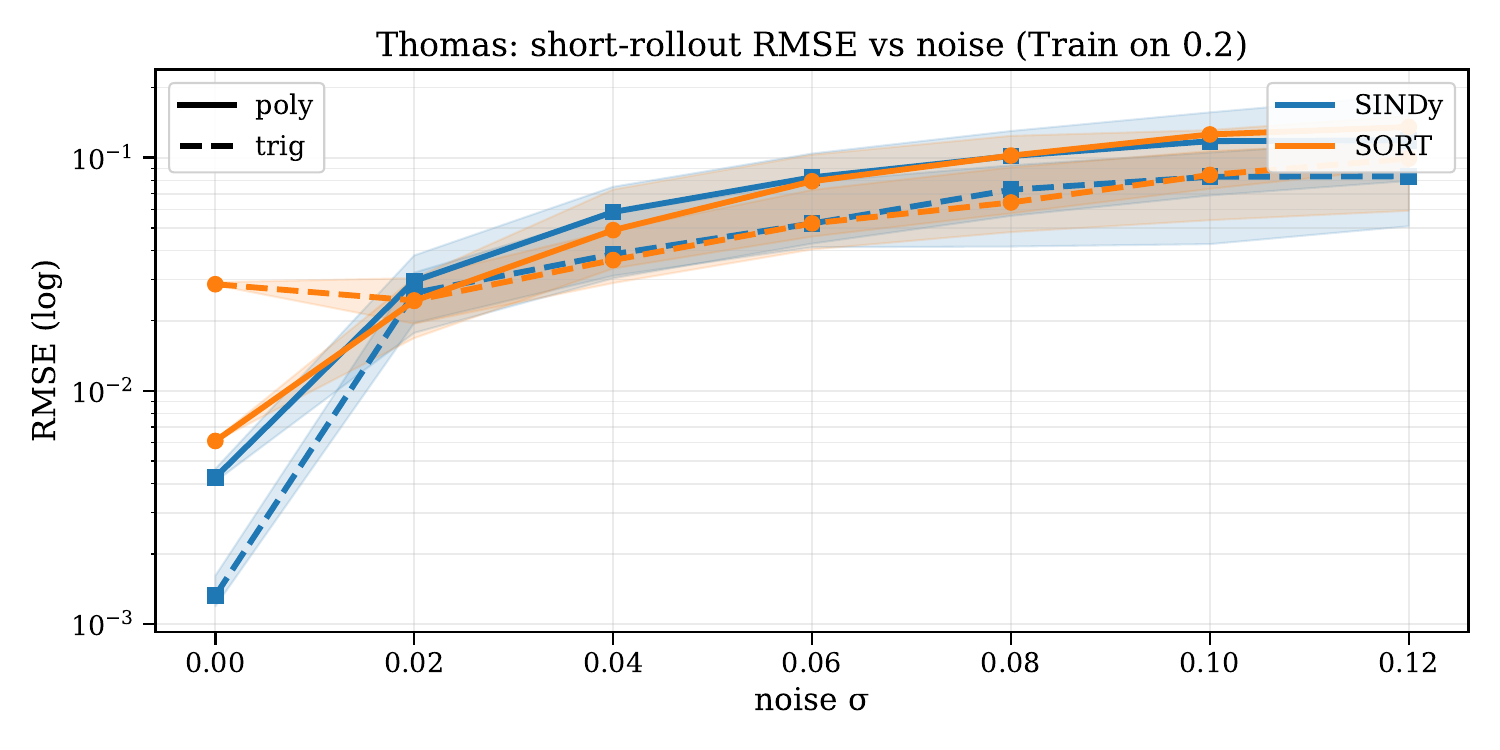}
    \label{fig:thomas_noise}
  }\\[2mm]
  \subfloat[BesselJ1: RMSE vs train fraction.]{
    \includegraphics[width=0.45\linewidth]{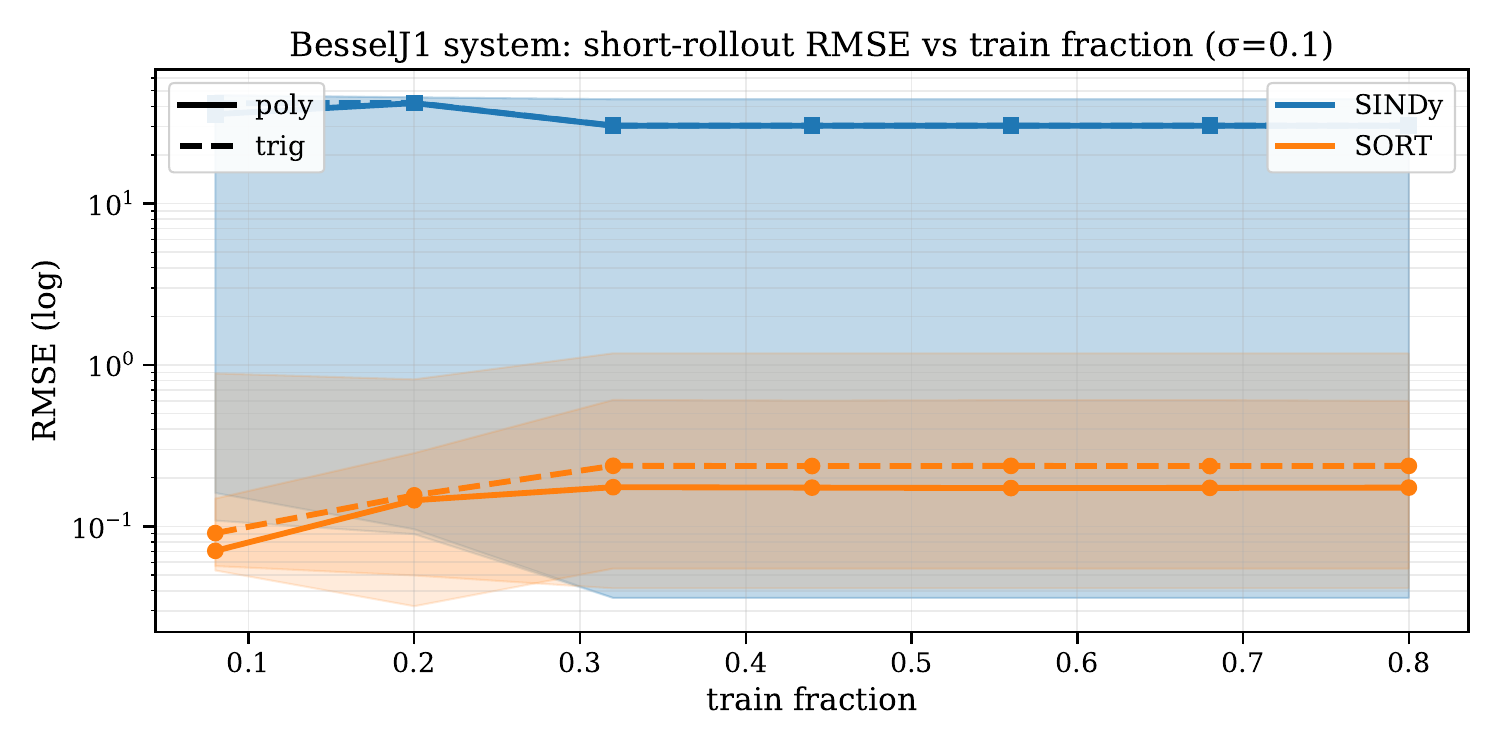}
    \label{fig:bessel_frac}
  }\hfill
  \subfloat[BesselJ1: RMSE vs noise.]{
    \includegraphics[width=0.45\linewidth]{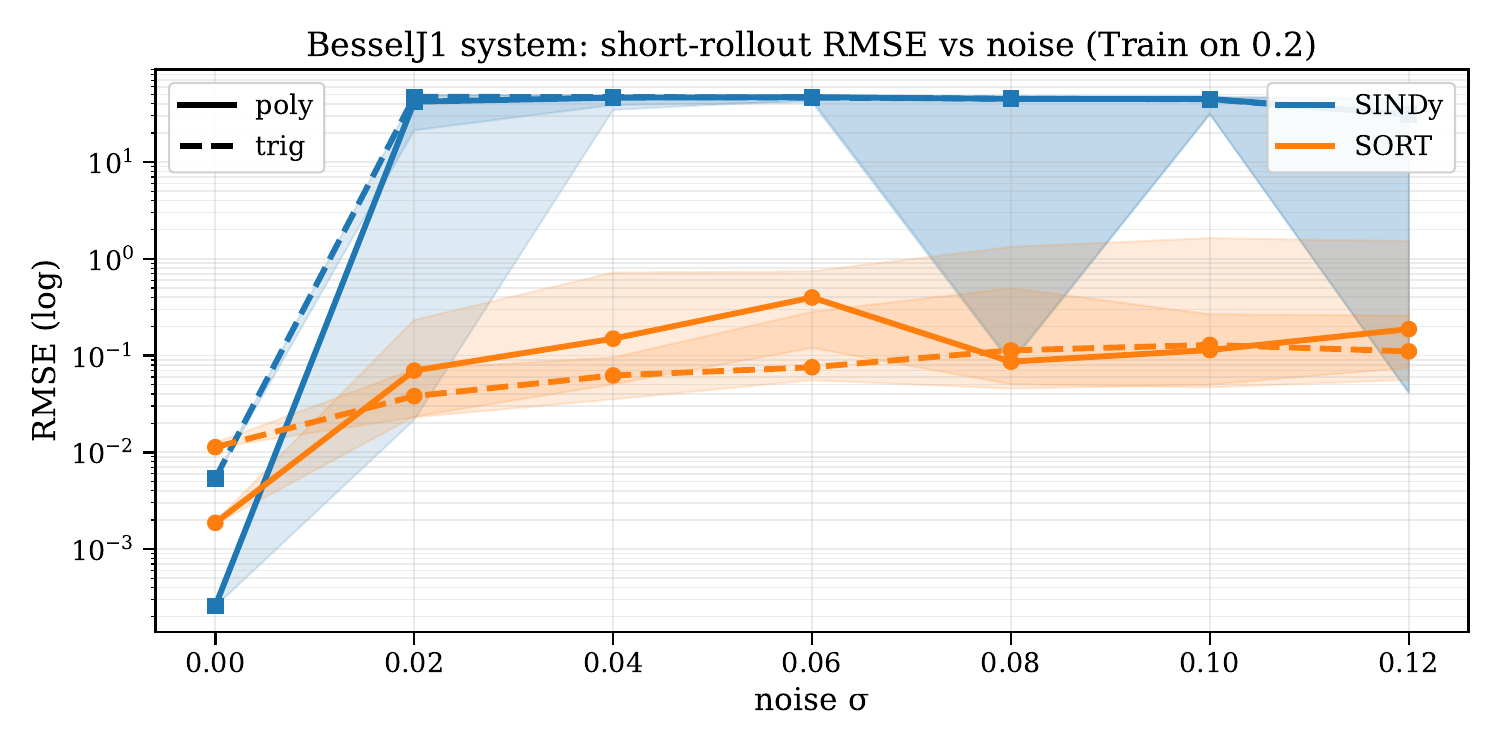}
    \label{fig:bessel_noise}
  }
  \caption{\textbf{Short-rollout prediction error for SINDy and SORT under basis choice.}
  Each panel reports short-horizon rollout RMSE on a log scale, shown as the median with interquartile-range band.
  Solid lines correspond to polynomial or Legendre bases; dashed lines correspond to trigonometric variants.
  For SINDy, dashed curves use a polynomial library augmented with trigonometric terms; for SORT, dashed curves use a trigonometric tensor basis.
  Panels (a)--(b) show the Thomas attractor, where both methods perform comparably and trigonometric features improve accuracy.
  Panels (c)--(d) show the Bessel-driven system with cyclic coupling through \(J_1(\cdot)\), where SORT remains robust across training fractions and noise levels while SINDy degrades under representation mismatch.
  Panels (a) and (c) vary the training fraction at fixed noise \(\sigma=0.1\); panels (b) and (d) vary the noise level at fixed training fraction \(20\%\).}
  \label{fig:thomas_bessel_panel}
\end{figure}

We consider two three-dimensional cyclic systems. The first is the Thomas attractor, whose dynamics are explicitly trigonometric. The second replaces the trigonometric coupling with a Bessel function \(J_1(\cdot)\):
\begin{center}
\begin{minipage}{0.48\textwidth}
\centering
\textbf{Thomas attractor}
\[
\begin{aligned}
\dot{x}_1 &= \sin(x_2) - b x_1,\\
\dot{x}_2 &= \sin(x_3) - b x_2,\\
\dot{x}_3 &= \sin(x_1) - b x_3 .
\end{aligned}
\]
\end{minipage}
\hfill
\begin{minipage}{0.48\textwidth}
\centering
\textbf{Bessel-driven variant}
\[
\begin{aligned}
\dot{x}_1 &= J_1(x_2) - b x_1,\\
\dot{x}_2 &= J_1(x_3) - b x_2,\\
\dot{x}_3 &= J_1(x_1) - b x_3 .
\end{aligned}
\]
\end{minipage}
\end{center}

Here \(J_1(x)\) denotes the Bessel function of the first kind of order one,
\[
\textstyle
J_1(x)=
\sum_{k=0}^{\infty}
\frac{(-1)^k}{k!(k+1)!}
\left(\frac{x}{2}\right)^{2k+1}
=
\frac{x}{2}-\frac{x^3}{16}+\frac{x^5}{384}-\cdots .
\]
This system is deliberately chosen to test representation mismatch: \(J_1(\cdot)\) is not sparse in low-degree polynomial or trigonometric libraries. The SINDy baseline uses either a polynomial library or a polynomial library augmented with trigonometric terms, while SORT uses either Legendre or trigonometric tensor bases with sparse regression.

Figure~\ref{fig:thomas_bessel_panel} summarizes the cyclic-system results using median short-rollout RMSE with interquartile ranges across random seeds. On the Thomas attractor, both methods perform comparably when the representation is well matched, particularly with trigonometric features. This is expected: the true dynamics align with the candidate representation, so the main differences are due to numerical conditioning and noise sensitivity.

The contrast is stronger for the Bessel-driven system. Here the SINDy-style baseline degrades as the training fraction decreases or the noise level increases, both with a polynomial library and with additional trigonometric terms. The issue is representation mismatch: the true nonlinearity is not sparse in the selected library, making coefficient selection brittle under noise. SORT remains more robust because it does not require the correct analytic building blocks to be anticipated in a finite dictionary. Instead, the vector field is approximated in orthonormal coordinates, where sparsity suppresses unstable modes and orthogonality reduces feature collinearity.

The benchmark study shows that, under sparse temporal sampling, sparse orthogonal regression can degrade more gradually than a SINDy-style library model using the same sampled states and finite-difference derivatives. The Thomas and Bessel experiments test noise, smoothing, and representation mismatch, showing that orthonormal expansions remain useful when the natural basis is uncertain or poorly captured by a small fixed library. Overall, the results support the view that basis choice is part of the modeling problem, not an implementation detail.

\subsection{Integral estimation}
\label{sec:integral-estimation}

Once an orthonormal expansion is learned, integrals can be evaluated as linear functionals of the estimated coefficients. SORT uses this property in a data-driven way: instead of computing expansion coefficients by quadrature or exact inner products, it estimates them from pointwise samples and then evaluates the integral by coefficient readout.

We work on a canonical domain \(t\in[-1,1]^d\), related to a bounded physical domain \(x\in[a,b]^d\) by the affine map
\begin{equation}
t_j=\frac{2x_j-(a+b)}{b-a},
\qquad
x_j=\frac{b-a}{2}\,t_j+\frac{a+b}{2}.
\label{eq:affine-map}
\end{equation}
Let
\[
D(t)=\prod_{j=1}^d (1+t_j^2),
\]
and define the transformed target
\[
g(t)=\frac{f(x(t))}{D(t)}.
\]
We expand \(g\) in a tensor-product orthonormal basis \(\{\Phi_\alpha\}_{\alpha\in\Lambda}\) on \([-1,1]^d\), choosing the basis element \(\Phi_{\mathbf{0}}\) as the normalized Riesz representer of the integration functional,
\[
\Phi_{\mathbf{0}}(t)=\frac{D(t)}{\|D\|_{L^2([-1,1]^d)}}.
\]
SORT fits a sparse expansion
\[
g(t)\approx\sum_{\alpha\in\Lambda_N} c_\alpha\,\Phi_\alpha(t).
\]
With this construction, integral estimation reduces to a single-coefficient readout:
\begin{equation}
\int_{[a,b]^d} f(x)\,dx
\;\approx\;
\Big(\frac{b-a}{2}\Big)^d\,
c_{\mathbf{0}}\,
\|D\|_{L^2([-1,1]^d)}.
\label{eq:integral-readout}
\end{equation}
Because \(D(t)\) factorizes across dimensions, its norm has a closed-form expression. Improper integrals can be handled either by mapping the unbounded domain to \([-1,1]^d\) or by reporting controlled finite-window approximations. In higher dimensions, sparse index sets such as hyperbolic-cross truncations are used to limit basis size and preserve sample efficiency.

\begin{figure*}[t]
    \centering
    \begin{subfigure}[t]{0.48\textwidth}
        \centering
        \includegraphics[width=\linewidth]{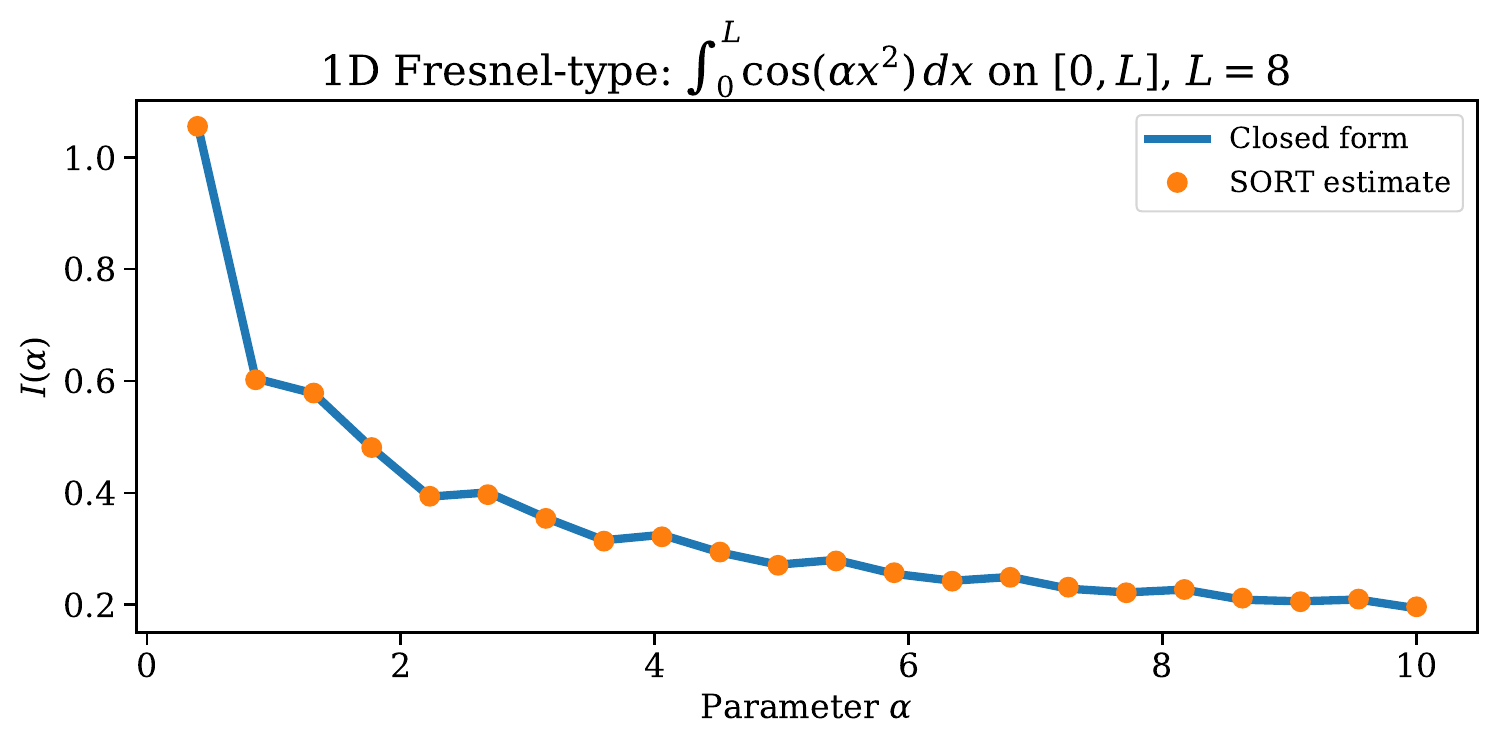}
        \caption{1D Fresnel-type integral}
        \label{fig:integral-fresnel-1d}
    \end{subfigure}
    \hfill
    \begin{subfigure}[t]{0.48\textwidth}
        \centering
        \includegraphics[width=\linewidth]{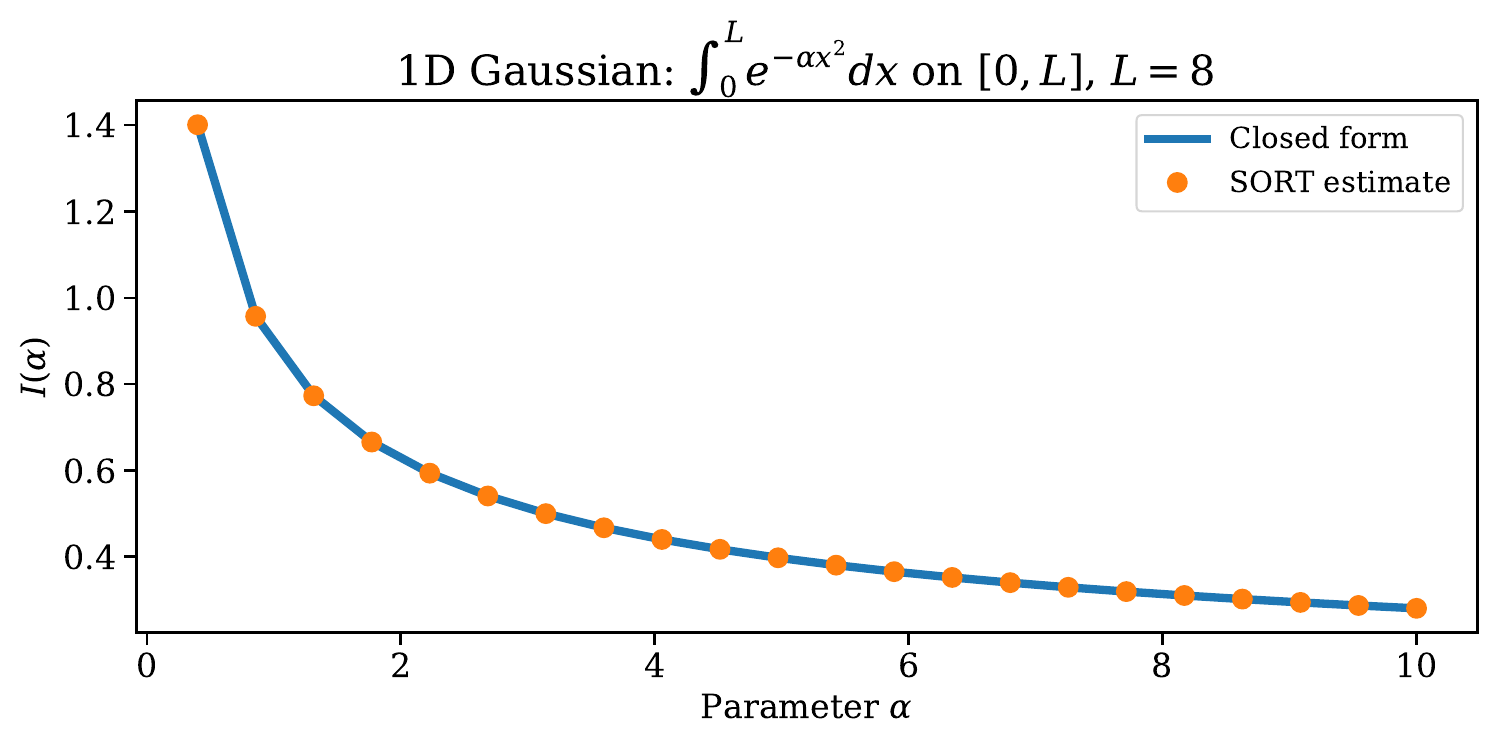}
        \caption{1D Gaussian integral}
        \label{fig:integral-gaussian-1d}
    \end{subfigure}\\
    \begin{subfigure}[t]{0.48\textwidth}
        \centering
        \includegraphics[width=\linewidth]{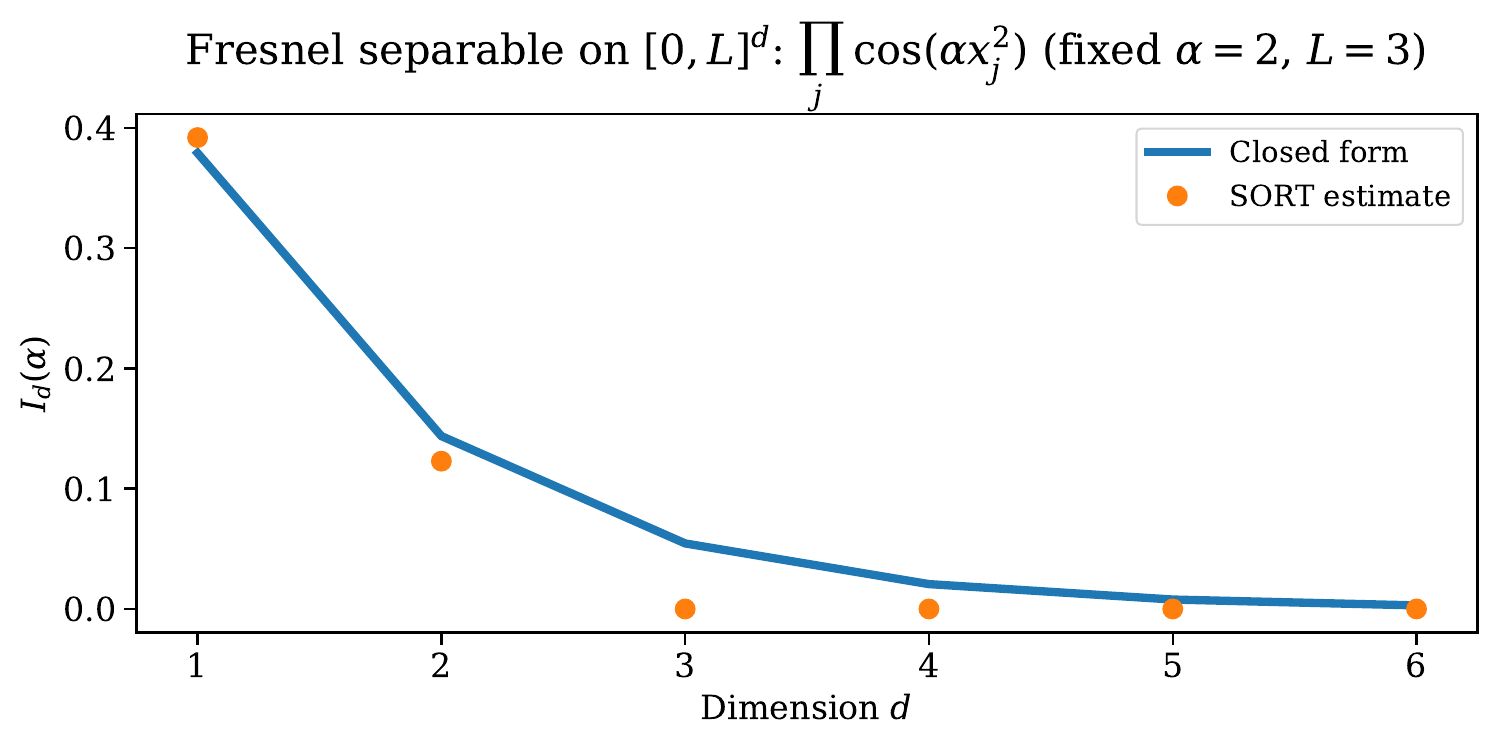}
        \caption{High-dimensional Fresnel integral}
        \label{fig:integral-fresnel-hd}
    \end{subfigure}
    \hfill
    \begin{subfigure}[t]{0.48\textwidth}
        \centering
        \includegraphics[width=\linewidth]{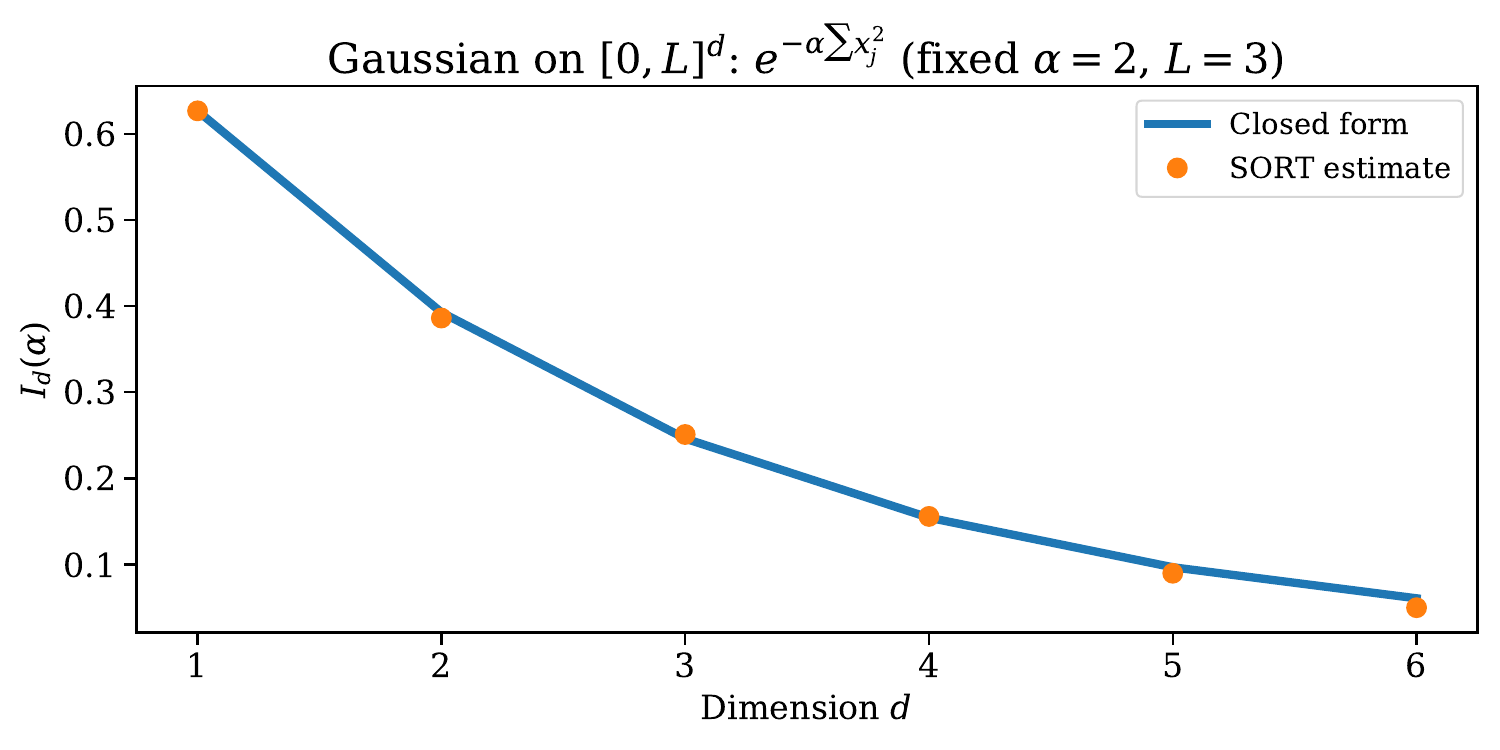}
        \caption{High-dimensional Gaussian integral}
        \label{fig:integral-gaussian-hd}
    \end{subfigure}
\caption{
\textbf{Integral estimation via SORT.}
SORT learns an orthonormal expansion of a transformed integrand from pointwise samples and estimates integrals by reading off the coefficient associated with the integration functional.
In one dimension, SORT matches closed-form values for both oscillatory Fresnel-type integrals and smooth Gaussian integrals across a range of parameters.
In higher dimensions, separable test integrands are used so that exact reference values remain available; accuracy decreases gradually with dimension, consistent with the increased sample complexity of sparse polynomial recovery.
}
    \label{fig:sor_integrals}
\end{figure*}

Figure~\ref{fig:sor_integrals} reports results for smooth and oscillatory integrands. Panels~(a)--(b) show one-dimensional examples: Gaussian integrals \(\int_0^L e^{-\alpha x^2}\,dx\) and Fresnel-type integrals \(\int_0^L \cos(\alpha x^2)\,dx\). SORT accurately tracks the closed-form values across a wide range of \(\alpha\), including strongly oscillatory Fresnel cases where accurate quadrature would require resolving rapid oscillations. Panels~(c)--(d) extend the evaluation to higher-dimensional separable integrals. The estimates remain accurate in moderate dimensions, but the error increases with dimension, reflecting the growth of the basis and the sample complexity of sparse coefficient recovery. Overall, these experiments illustrate the second application-level contribution of SORT: once a sparse expansion is recoverable, integration becomes a stable coefficient-readout task rather than a separate quadrature problem.

\begin{figure*}[]
    \centering
    \begin{subfigure}[t]{0.48\textwidth}
        \centering
        \includegraphics[width=\linewidth]{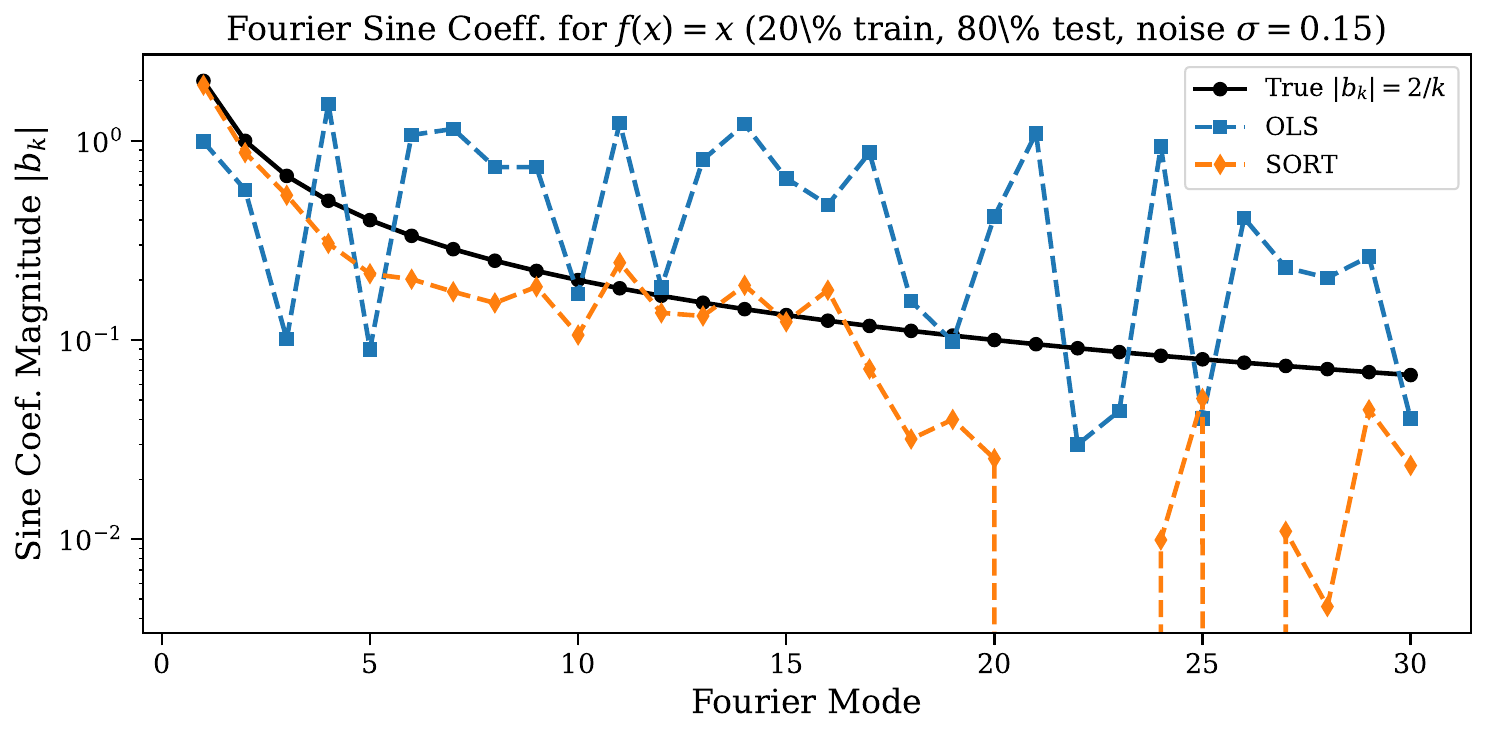}
        \caption{Fourier coefficients for \(f(x)=x\)}
        \label{fig:fourier-coeffs}
    \end{subfigure}
    \hfill
    \begin{subfigure}[t]{0.48\textwidth}
        \centering
        \includegraphics[width=\linewidth]{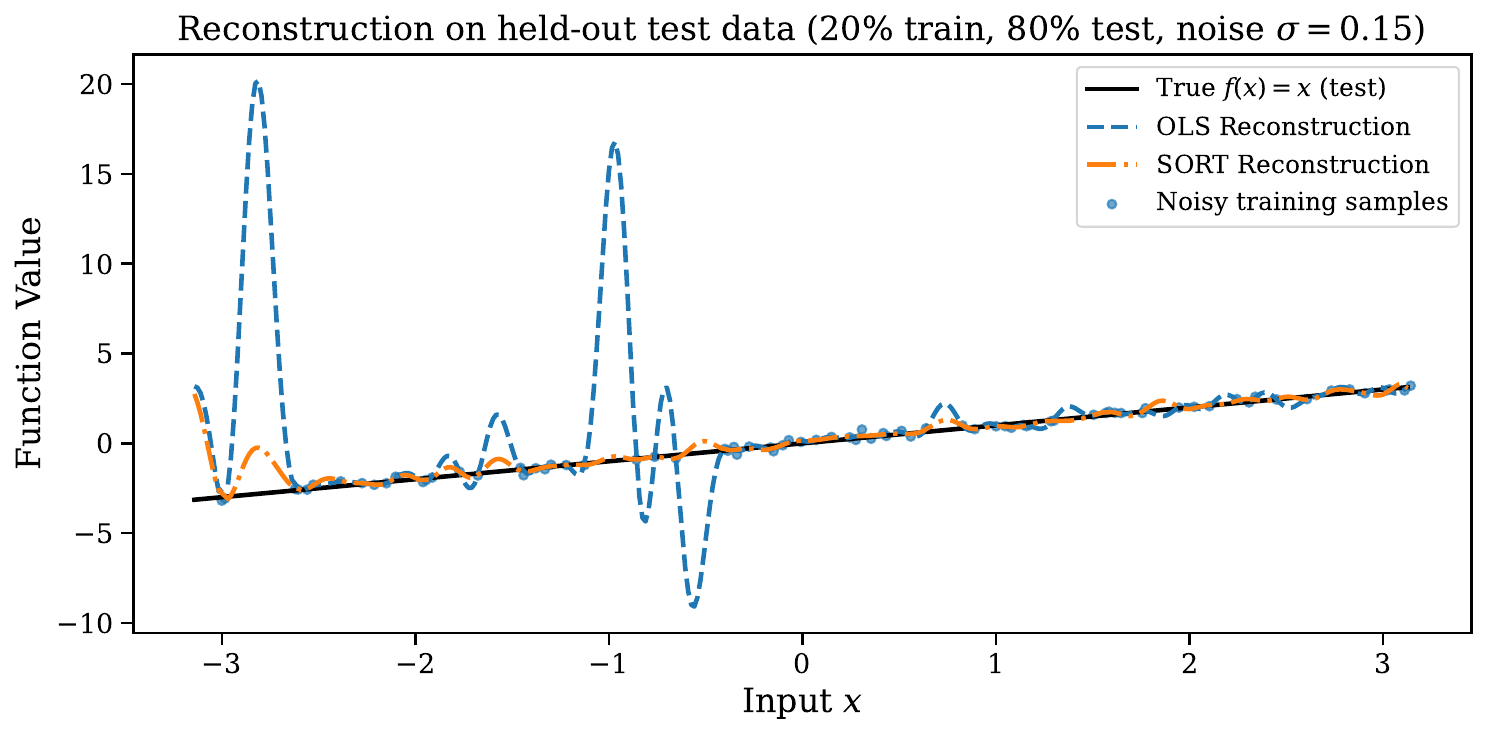}
        \caption{Held-out reconstruction}
        \label{fig:fourier-reconstruction}
    \end{subfigure}\\[-1mm]
    \begin{subfigure}[t]{0.48\textwidth}
        \centering
        \includegraphics[width=\linewidth]{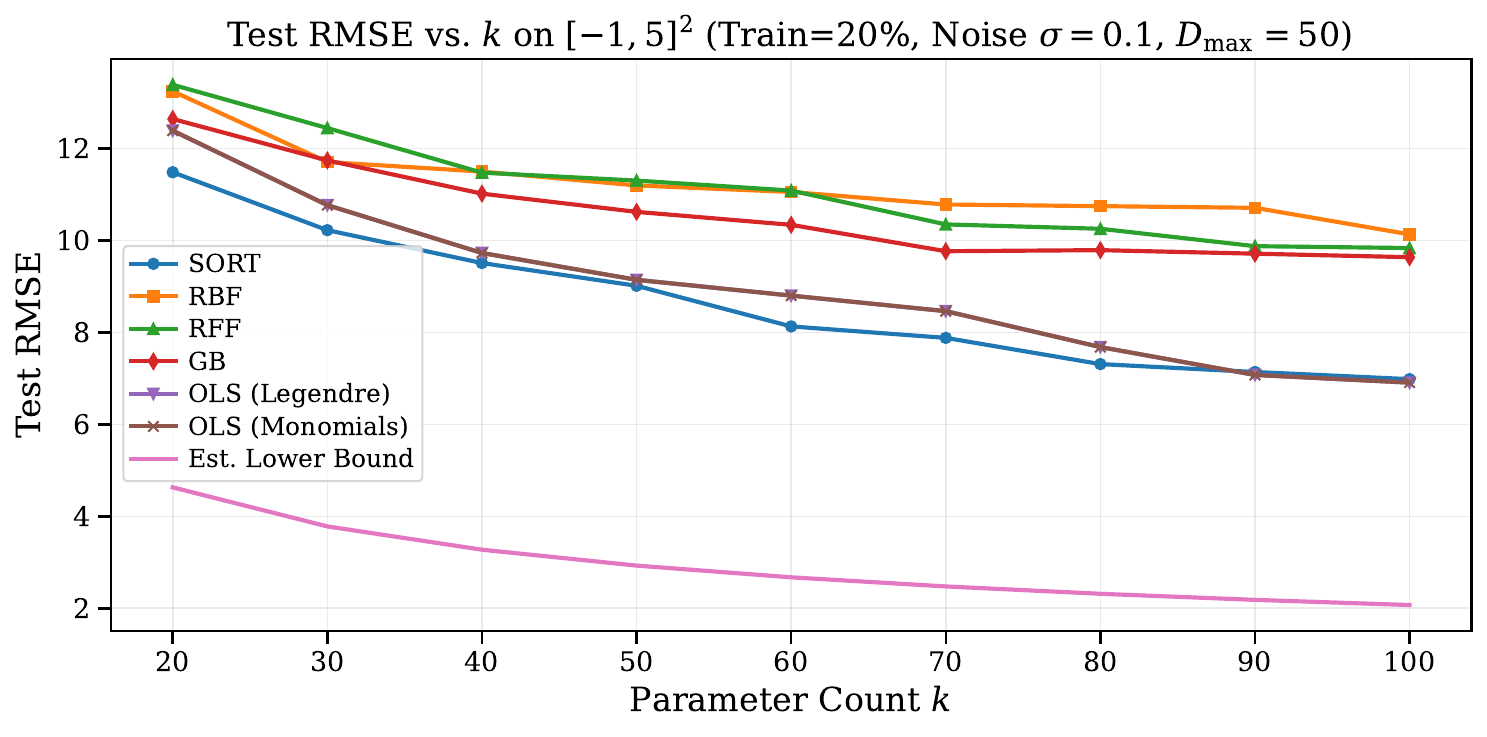}
        \caption{Test RMSE vs parameter count}
        \label{fig:rmse-vs-k}
    \end{subfigure}
    \hfill
    \begin{subfigure}[t]{0.48\textwidth}
        \centering
        \includegraphics[width=\linewidth]{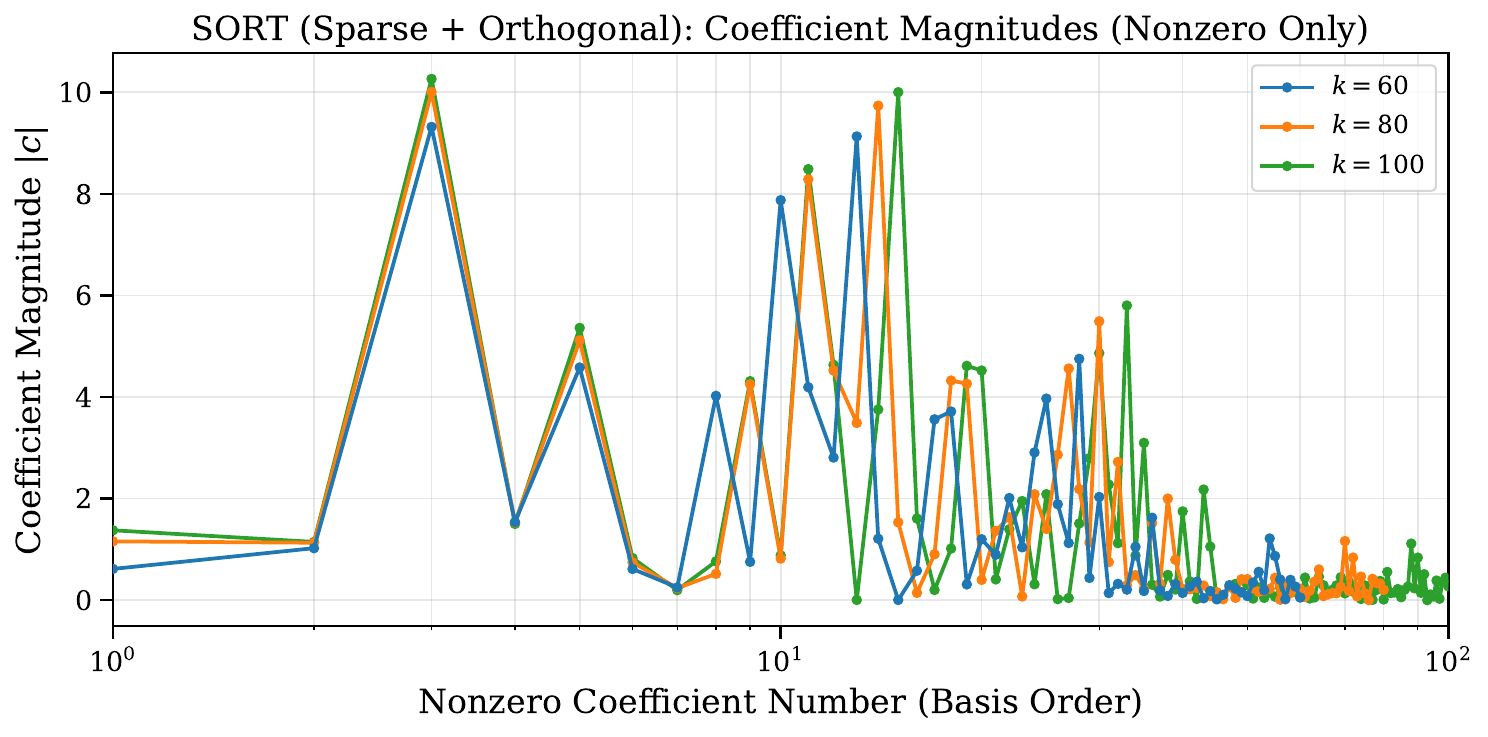}
        \caption{SORT coefficients}
        \label{fig:sor-coeffs}
    \end{subfigure}\\[-1mm]
    \begin{subfigure}[t]{0.48\textwidth}
        \centering
        \includegraphics[width=\linewidth]{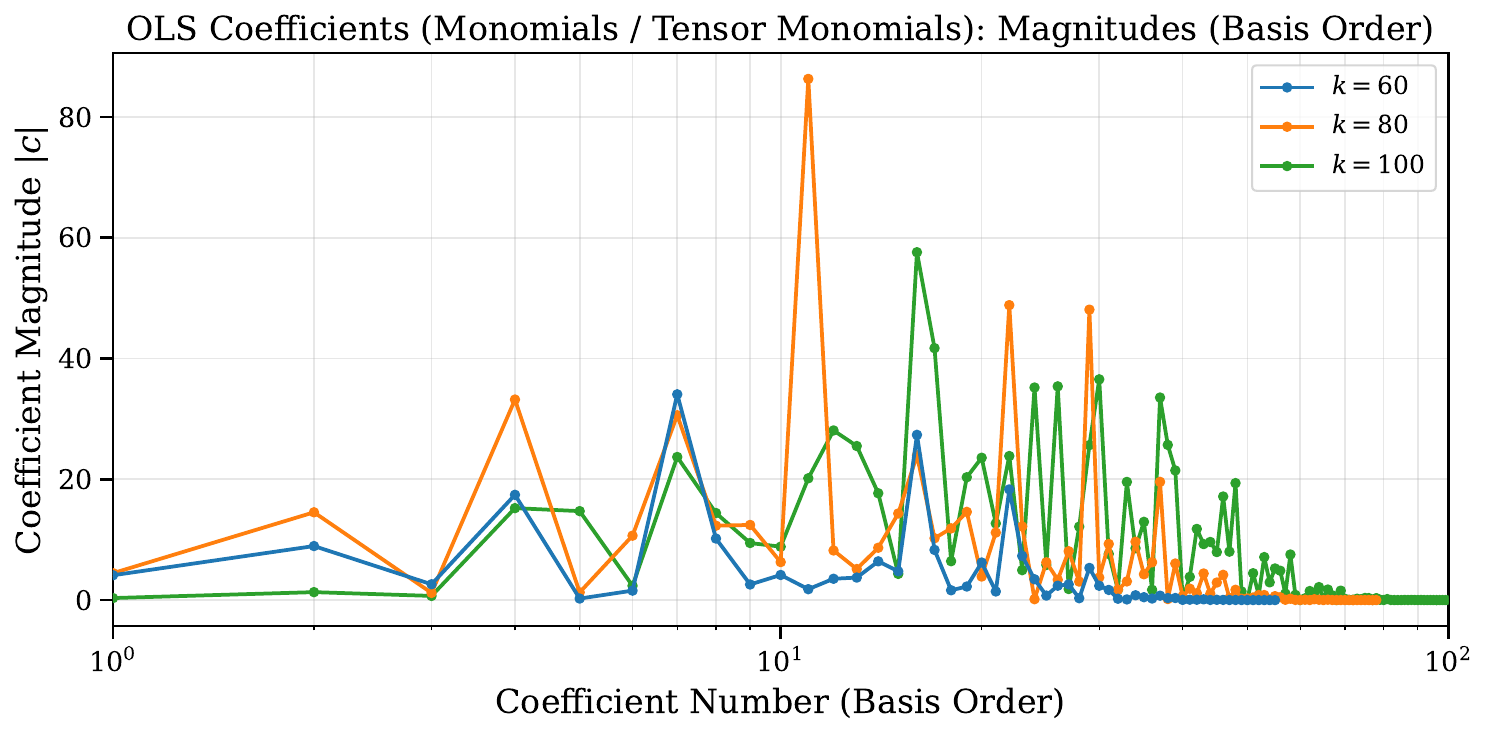}
        \caption{OLS coefficients, monomials}
        \label{fig:ols-monomial-coeffs}
    \end{subfigure}
    \hfill
    \begin{subfigure}[t]{0.48\textwidth}
        \centering
        \includegraphics[width=\linewidth]{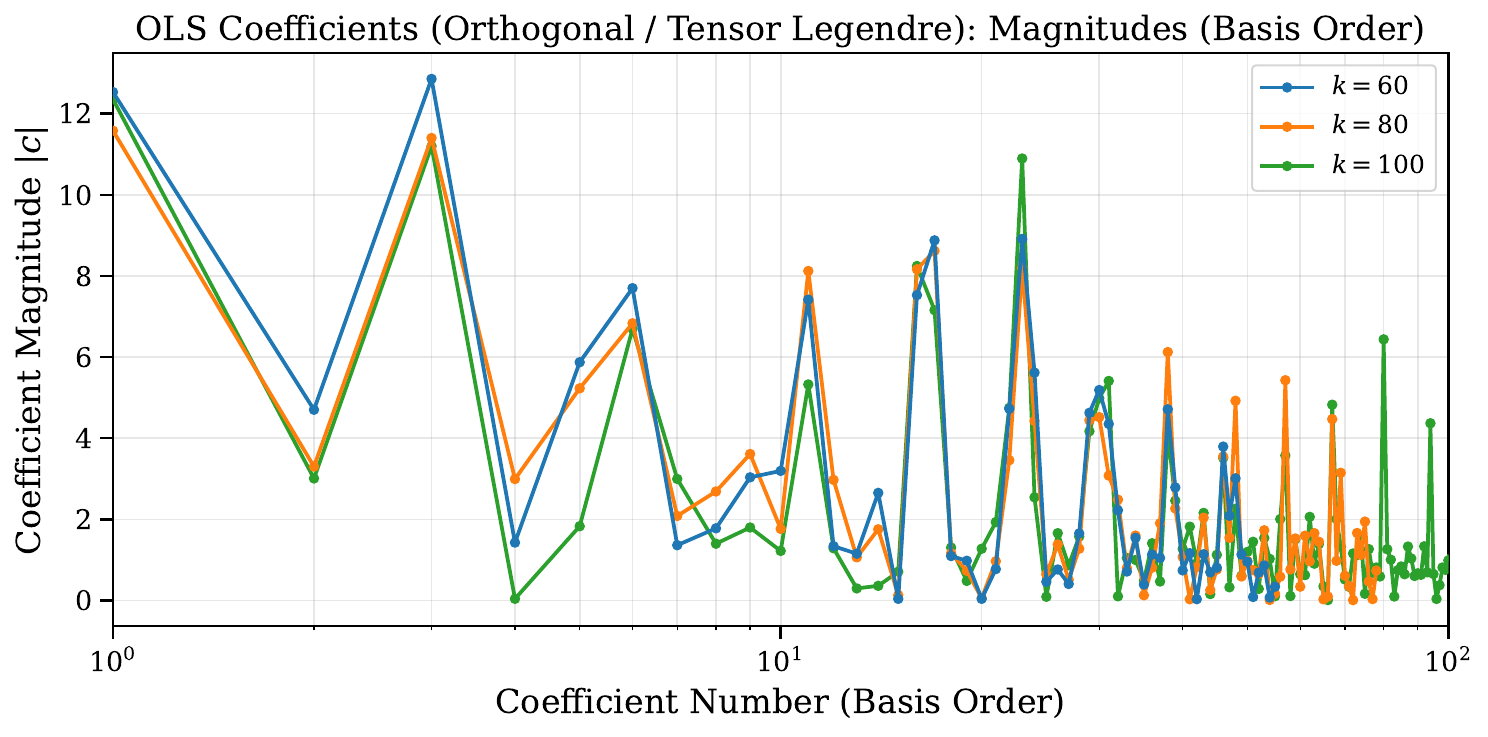}
        \caption{OLS coefficients, Legendre}
        \label{fig:ols-legendre-coeffs}
    \end{subfigure}\\[-1mm]
    \begin{subfigure}[t]{0.48\textwidth}
        \centering
        \includegraphics[width=\linewidth]{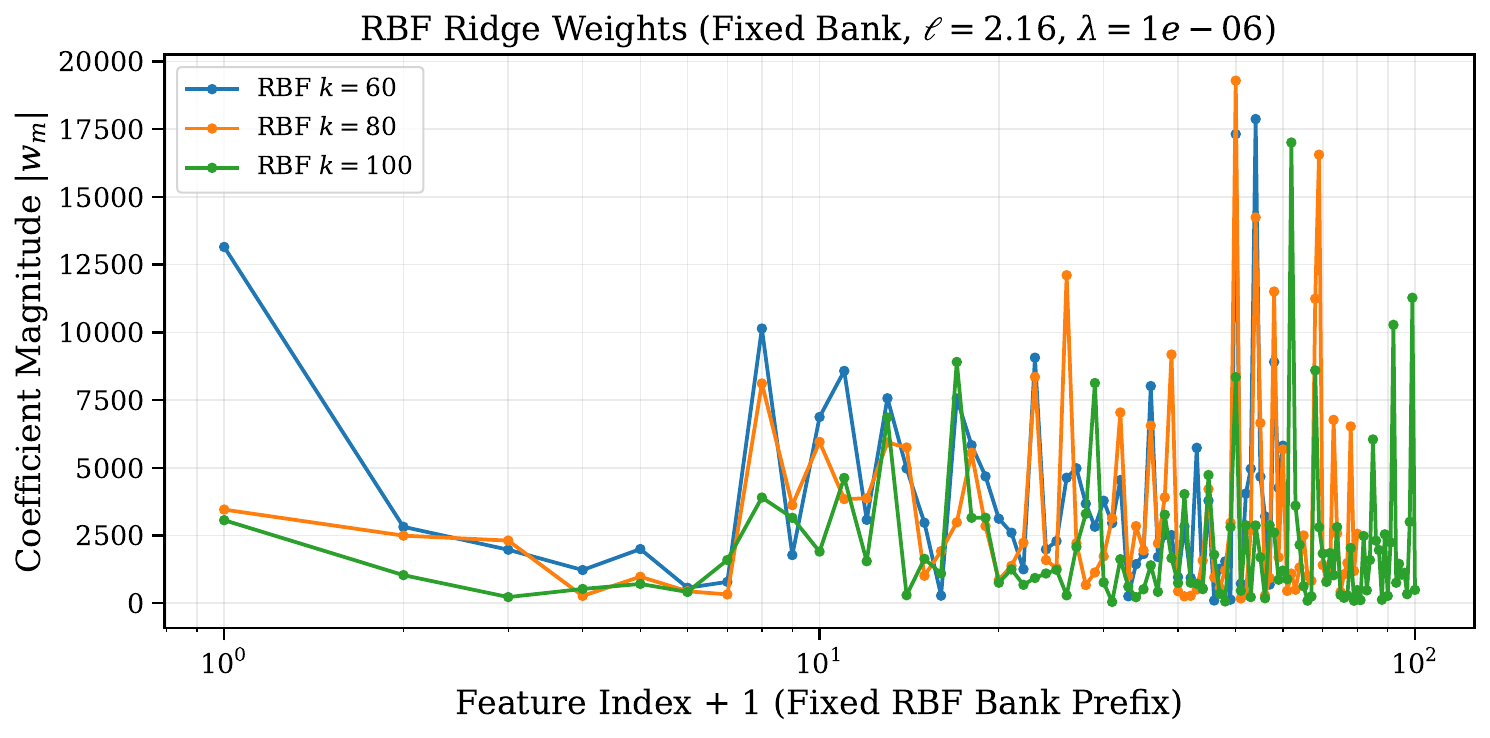}
        \caption{RBF ridge coefficients}
        \label{fig:rbf-coeffs}
    \end{subfigure}
    \hfill
    \begin{subfigure}[t]{0.48\textwidth}
        \centering
        \includegraphics[width=\linewidth]{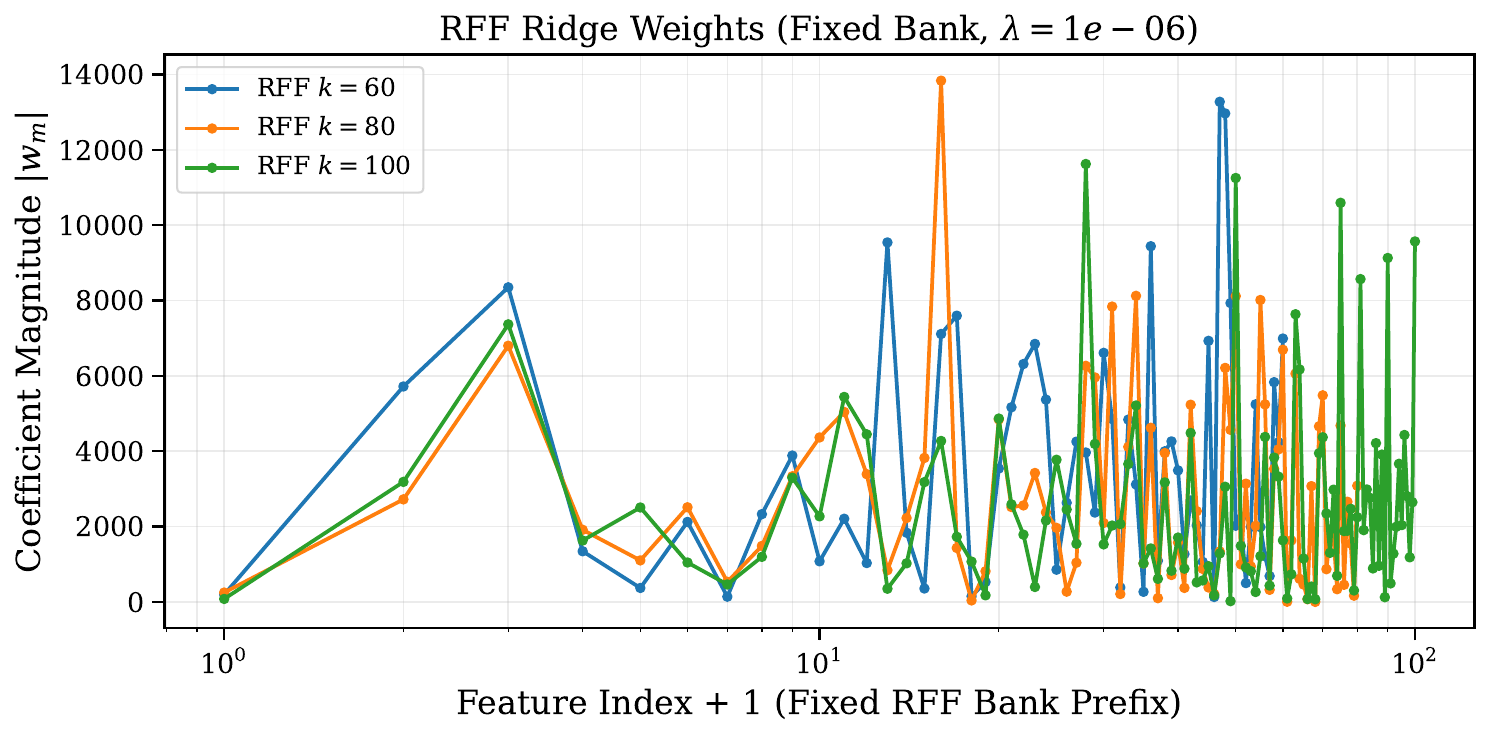}
        \caption{RFF ridge coefficients}
        \label{fig:rff-coeffs}
    \end{subfigure}
\caption{
\textbf{Spectral recovery and order-consistent approximation.}
(a)--(b) Recovery of the Fourier sine structure of \(f(x)=x\) on \([-\pi,\pi]\) from noisy data using \(20\%\) of the samples for training and \(80\%\) for testing. OLS amplifies unstable high-frequency modes, while SORT recovers the dominant \(1/k\) coefficient decay and gives a stable reconstruction.
(c)--(h) Approximation of \(f(x_1,x_2)=\exp(x_1\cos(2x_2))\) on \([-1,5]^2\) using \(20\%\) noisy training data \((\sigma=0.1)\) and \(80\%\) clean test data. SORT achieves low test error and a stable low-order coefficient hierarchy, while dense monomial, kernel, and random-feature representations produce coefficients that vary more strongly with model size.
}
    \label{fig:spectral_order_consistency}
\end{figure*}

\subsection{Spectral recovery and order-consistent approximation}
\label{sec:spectral-recovery}

We finally use two controlled approximation examples to illustrate a standard but useful property of orthonormal expansions: increasing the model order refines the representation without changing the meaning of previously learned low-order coefficients. This is not a new fact from approximation theory. The point is that, when combined with sparsity, it gives a practical machine-learning diagnostic for stable model growth under noise and subsampling.

The first example has known spectral structure: \(f(x)=x\) on \([-\pi,\pi]\) has Fourier sine coefficients \(|b_k|=2/k\). Figure~\ref{fig:spectral_order_consistency}(a)--(b) shows that, under noise and subsampling, OLS fits unstable high-frequency coefficients that lead to oscillatory test error. SORT instead suppresses unsupported modes and preserves the dominant low-order spectral structure.

The second example considers
\[
f(x_1,x_2)=\exp\!\big(x_1\cos(2x_2)\big), \qquad (x_1,x_2)\in[-1,5]^2,
\]
using \(20\%\) noisy training data, \(80\%\) clean test data, tensor-product Legendre bases with total-degree cutoff \(D_{\max}=50\), and parameter counts \(k\in\{60,80,100\}\). Figure~\ref{fig:spectral_order_consistency}(c)--(h) compares SORT with dense least-squares, RBF, RFF, and gradient-boosting baselines. SORT gives the lowest test error and, more importantly, preserves a stable low-order coefficient hierarchy as model size increases. OLS in a monomial basis changes substantially with order; OLS in a Legendre basis benefits from orthogonality but remains noise-sensitive; and RBF/RFF ridge models produce dense weights without an interpretable hierarchy.

These examples illustrate the approximation-side contribution of SORT. Sparse regression in orthogonal bases is standard, but its order-consistent structure is underused as a machine-learning design principle: model complexity can be increased by adding higher-order terms while retaining the interpretation of previously learned low-order coefficients. This is also the sense in which the name SORT is descriptive: models can be compared across truncation orders by their active coefficient sets, coefficient persistence, sparsity level, and predictive accuracy, because all models are expressed in the same ordered orthonormal coordinate system.

\section{Conclusion}
\label{sec:conclusion}

This paper developed SORT as a sparse spectral representation framework built from orthonormal basis expansions and sparsity-promoting regression. The contribution is application-level: SORT turns a learned coefficient vector into a reusable object for integration, dynamical-system reconstruction, and nonlinear approximation within a common representation.

The dynamical-system experiments show that sparse orthonormal vector-field representations are useful when derivative information is degraded or when the natural basis is uncertain. In the sampling-degradation study, SORT remains competitive with SINDy-style sparse identification at fine sampling and degrades less abruptly under coarse sampling. In the cyclic-system experiments, it performs comparably on the Thomas attractor when trigonometric structure is supplied to both methods, and is more robust on the Bessel-driven variant, where the dynamics are poorly captured by a small polynomial or trigonometric library. These results support the view that basis design is part of the scientific modeling problem, not an implementation detail.

The integration experiments show that data-driven quadrature can be treated as coefficient readout. Rather than constructing explicit quadrature rules or computing inner products analytically, SORT estimates expansion coefficients from samples and evaluates the integral through the coefficient associated with the integration functional. This works for smooth Gaussian integrals, oscillatory Fresnel-type integrals, and moderate-dimensional separable examples when the expansion remains sparse.

The approximation experiments highlight the value of order-consistent model growth. Because orthonormal expansions provide a stable hierarchy of basis functions, increasing the model order can refine the representation without changing the meaning of lower-order coefficients. In noisy finite data, coefficient persistence across truncation levels gives a practical diagnostic for stable learning.

Overall, SORT provides a compact intermediate representation between raw samples and later symbolic or analytic description. It can first recover a stable expansion in a suitable orthogonal basis; a subsequent symbolic stage may then search for simpler closed-form expressions consistent with that expansion. Future work should develop adaptive and anisotropic basis selection, empirical orthogonalization for irregular samples, and larger-scale applications. The results here suggest that sparse orthonormal expansions learned from data offer a stable, reusable, and computationally simple route to equation discovery, integration, and approximation.

\section*{Code Availability}
\vspace{-0.2cm}
Code for reproducing the experiments and figures is available at:\\
\url{https://doi.org/10.5281/zenodo.21707070}.
\vspace{-0.2cm}
\begin{credits}
\subsection*{Disclaimer}
Co-funded by the European Union. Views and opinions expressed are however those of the author(s) only and do not necessarily reflect those of the European Union or European Research Executive Agency. Neither the European Union nor the granting authority can be held responsible for them.
\vspace{-0.4cm}
\subsection*{\ackname}
This publication is supported by the European Union's Horizon Europe research and innovation programme under the Marie Sk\l{}odowska-Curie Postdoctoral Fellowship Programme, SMASH co-funded under the grant agreement No. 101081355. The operation (SMASH project) is co-funded by the Republic of Slovenia and the European Union from the European Regional Development Fund.

The authors acknowledge the financial support of the Slovenian Research Agency via the Gravity project \textsl{AI for Science}, GC-0001 and of the Slovenian Research And Innovation Agency (research core funding No. P1-0188).
\end{credits}
\vspace{-0.4cm}
%
%
%
\bibliographystyle{splncs04}
\bibliography{biblio}

\end{document}